\documentclass[10pt,twocolumn,letterpaper]{article}

\pdfoutput=1

\usepackage{graphicx}
\usepackage[pdftex]{color}
\usepackage{tabularx}
\usepackage{adjustbox}
\usepackage{multirow}
\usepackage{comment}
\usepackage{array}
\usepackage{pifont} 
\usepackage{longtable}
\usepackage{authblk}

\usepackage{cvpr}              

\definecolor{cvprblue}{rgb}{0.21,0.49,0.74}
\usepackage[pagebackref,colorlinks,allcolors=cvprblue]{hyperref}

\newcommand{\tbd}[1]{{\color{magenta}{#1}}}
\newcommand{\kt}[1]{{\color{black}{#1}}}

\def\paperID{2148} 
\def\confName{CVPR}
\def\confYear{2025}

\title{
Text-guided Synthetic Geometric Augmentation for Zero-shot 3D Understanding
}

\author[1,2]{Kohei Torimi}
\author[2,3]{Ryosuke Yamada}
\author[4]{Daichi Otsuka}
\author[2]{Kensho Hara}
\author[5]{\newline Yuki M. Asano}
\author[2,6]{Hirokatsu Kataoka}
\author[1]{Yoshimitsu Aoki}

\affil[1]{Keio University}
\affil[2]{National Institute of Advanced Industrial Science and Technology (AIST)}
\affil[3]{University of Tsukuba}
\affil[4]{TICO-AIST Cooperative Research Laboratory for Advanced Logistics (ALlab)}
\affil[5]{University of Technology Nuremberg}
\affil[6]{University of Oxford}

\affil[ ]{\tt\small{\{kohei.torimi, ryosuke.yamada, ootsuka.da, kensho.hara\}@aist.go.jp},}
\affil[ ]{\tt\small{yuki.asano@utn.de, hirokatsu.kataoka@aist.go.jp, aoki@elec.keio.ac.jp}}

\begin{document}
\maketitle
\begin{abstract}

Zero-shot recognition models require extensive training data for generalization.
However, in zero-shot 3D classification, collecting 3D data and captions is costly and labor-intensive, posing a significant barrier compared to 2D vision.
Recent advances in generative models have achieved unprecedented realism in synthetic data production, and recent research shows the potential for using generated data as training data.
Here, naturally raising the question: Can synthetic 3D data generated by generative models be used as expanding limited 3D datasets? 
In response, we present a synthetic 3D dataset expansion method, \textbf{Te}xt-guided \textbf{G}eometric \textbf{A}ugmentation (\textbf{TeGA}).
TeGA is tailored for language-image-3D pretraining, which achieves SoTA in zero-shot 3D classification, and uses a generative text-to-3D model to enhance and extend limited 3D datasets.
Specifically, we automatically generate text-guided synthetic 3D data and introduce a consistency filtering strategy to discard noisy samples where semantics and geometric shapes do not match with text. 
In the experiment to double the original dataset size using TeGA, our approach demonstrates improvements over the baselines, achieving zero-shot performance gains of 3.0\% on Objaverse-LVIS, 4.6\% on ScanObjectNN, and 8.7\% on ModelNet40.
These results demonstrate that TeGA effectively bridges the 3D data gap, enabling robust zero-shot 3D classification even with limited real training data and paving the way for zero-shot 3D vision applications. 

\end{abstract}

\section{Introduction}
\label{intro}

\begin{figure}[t]
    \centering
    \includegraphics[width=1.0\linewidth]{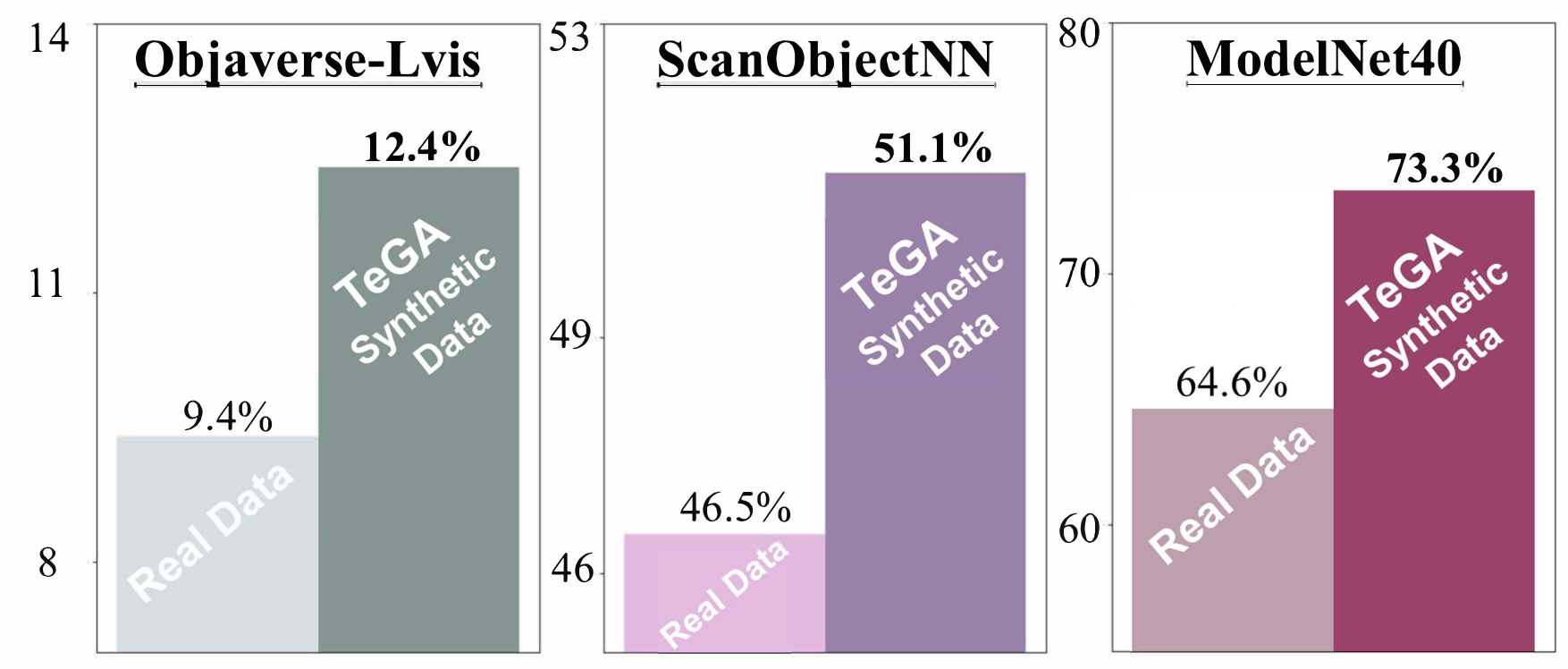}
    \caption{Our proposed TeGA (Text-guided Geometric Augmentation) assigns text guidance and a generative text-to-3D model for high-efficient dataset expansion which dramatically augments limited real data. Although we employ existing methods (\textit{e.g.,} Point-E) and simple tricks within text prompting, the proposal performs enough noteworthy results that 3D dataset with add synthetic data with TeGA outperforms ShapeNet trained model on e.g., Objaverse-LVIS, ModenNet-40 and ScanObjectNN under the setting of zero-shot 3D classification.
    }
    \label{fig1}
\end{figure}

Zero-shot recognition models have made remarkable progress leveraging paired image-text data.  
In particular, CLIP~\cite{radford2021learning} has shown the strong potential such models offer for zero-shot open-vocabulary image classification. Highly accurate zero-shot recognition models are increasingly applied in industrial applications too, such as robotics, manufacturing, and autonomous driving. Yet, this progress is so far mostly limited to the 2D domain. We argue that multi-modal representation learning that leverages not only images and text but also 3D data is critical for zero-shot 3d classification tasks.

One of the reasons for the success of vision-language models such as CLIP lies in the vast scale of training data utilized. These range from 400 million~\cite{radford2021learning} to several billions of image-text pairs collected from the Web~\cite{schuhmann2022laion}. 
However, collecting 3D data is much more challenging and costly due to the scarcity of high-quality 3D data on the Web. This is because the primary approaches require capturing 3D data in real-world environments with LiDAR or manually creating CAD models.
Consequently, zero-shot recognition in 3D vision has lagged behind progress in 2D vision, and the limited availability of 3D datasets remains a bottleneck issue. 
To address this challenge, recent research on zero-shot 3D classification mainly focuses on distilling knowledge from a large-scale pre-trained visual language model.
However, despite these efforts, the knowledge distillation does not fundamentally fix the issue of the limited availability of 3D data.


To this end, we propose to use recent text-to-3D generative models. 
These can produce highly realistic 3D synthetic data that are almost indistinguishable from the training data~\cite{liu2023zero,poole2022dreamfusion,lin2023magic3d,jun2023shap,nichol2022point}.
This remarkable evolution of the text-to-3D model simply begs the question: Can synthetic 3D data generated from text-to-3D models be used as expanding limited 3D training datasets? 


This paper thus proposes a dataset expansion method for zero-shot 3D recognition called \textbf{Te}xt-guided \textbf{G}eometric \textbf{A}ugmentation (\textbf{TeGA}). 
The proposed TeGA expands the training language-image-3D dataset for models that perform knowledge distillation from visual language models and enables reducing the costs associated with 3D data collection and annotation.
Specifically, we automatically generate synthetic 3D data and rendered images using an off-the-shelf text-to-3D model (\textit{e.g.,} Point-E~\cite{nichol2022point}) and use the latent space of the generative model to generate diverse geometric shapes with the same semantic content based on text prompts. 
Furthermore, TeGA also introduces a consistency filtering strategy to remove noisy data that does not match the text prompt, \textit{i.e.,} misalignment between language, images, and 3D data. 

To verify the effectiveness of TeGA, we train MixCon3D~\cite{gao2024sculpting} with ShapeNet and the synthetic dataset generated by TeGA and perform experiments on three representative zero-shot 3D classification datasets, Objaverse-LVIS~\cite{deitke2023objaverse}, ScanObjectNN~\cite{uy2019revisiting} and ModelNet40~\cite{wu20153d}
As a result, we show that TeGA improves performance in zero-shot 3D classification across three benchmark datasets. Specifically, TeGA achieves 12.4\% on Objaverse-LVIS, 51.1\% on ModelNet40, 73.3\% on ScanObjectNN, and  as shown in Figure~\ref{fig1}.
These results suggest that combining real and synthetic data improves the generalisation of visual models. We also found that consistency filtering plays important roles in learning. Our contributions in this paper are as follows:

\begin{itemize}
    \item We introduce TeGA, a method for dataset expansion that leverages text-guided generative models to tackle the problem of 3D data scarcity. TeGA enables us to automatically generate synthetic language-image-3D data tailored to specific semantics category of original 3D dataset by introducing text-guided prompt and consistency filtering.
    \item TeGA enhances zero-shot 3D classification by expanding existing datasets with synthetic 3D data generated from a generative model, addressing the challenges of data scarcity in 3D vision tasks.
\end{itemize}

\section{Related Work}
\noindent{\textbf{Multi-Modal Representation Learning.}} In zero-shot 3D classification, recent research has focused on multi-modal learning with limited 3D data by distilling CLIP knowledge~\cite{qi2023contrast, hegde2023clip, huang2023clip2point, xue2024ulip, liu2024openshape} because CLIP possesses extensive knowledge in vision-language tasks. For example, MixCon3D~\cite{gao2024sculpting} improved the performance of zero-shot 3D classification by contrastive learning across the image-text-point cloud using CLIP knowledge. 
In addition, ULIP-2~\cite{xue2024ulip} is a multimodal pre-training that automatically generates comprehensive language descriptions for 3D shapes without manual annotation.
These conventional methods achieve high zero-shot performance on ModelNet40~\cite{wu20153d}, ScanObjectNN~\cite{uy2019revisiting}, and Objaverse-LVIS~\cite{deitke2023objaverse}. However, these methods basically focus only on improving training methods, and there are limits to how much performance can be improved with this approach alone. We believe it is essential for the model and data sides to evolve together. This paper thus explores the potential of synthetic dataset expansion to achieve scalable geometric learning for zero-shot 3D classification.

\noindent{\textbf{Text-to-3D Models.}} Recent text-to-image models have experienced rapid growth. Inspired by this, text-to-3D models have also become a prominent area of research. 
Text-to-3D models also face the challenge of limited 3D data, making it difficult to generate open-vocabulary 3D data directly from text. To address this, recent text-to-3D models employ a strategy that leverages text-to-2D models or CLIP's extensive language and image knowledge. They first generate images or image features and then lift these images to 3D data~\cite{poole2022dreamfusion, lin2023magic3d, chen2023fantasia3d, wang2024prolificdreamer, tsalicoglou2024textmesh}.
For example, DreamFusion~\cite{poole2022dreamfusion} efficiently generates 3D data using CLIP's knowledge and NeRF's gradient updates.
In addition, Zero123-XL~\cite{liu2023zero} trains on the relatively large 3D dataset called Objaverse-XL~\cite{deitke2024objaverse} and achieves highly accurate 3D generation. Recently, it has also been applied to explicit 3D representations such as point clouds and meshes~\cite{nichol2022point, jun2023shap}. Using a text-to-image and image-to-point cloud pipeline, Point-E~\cite{nichol2022point} quickly generates point clouds that correspond to text prompts.
In this paper, we utilize text-to-3D models for dataset expansion.

\noindent{\textbf{Generative Models as Data++.}} 
Recent generative models have evolved to generate realistic synthetic data that is visually almost indistinguishable~\cite{rombach2022high, saharia2022photorealistic, nichol2021glide}. And then the data generated from generative models can be thought of as data for recognition models with controllability and rich representation as data++~\cite{isola2022when, sariyildiz2023fake}. It reduces the collecting cost with real-world data while enabling the efficient generation of meaningful synthetic datasets. For example, StableRep~\cite{tian2024stablerep} learns visual representations by incorporating synthetic images generated by Stable Diffusion~\cite{rombach2022high} into multi-positive contrastive learning.  
StableRep is designed to learn images comprehensively across different views by treating multiple synthetic images of the same input text prompt as positive samples. 
In addition, SynCLR~\cite{tian2024learning} performs contrastive learning using only synthetic images and captions.
Despite training without real images, it performs equally well and better than traditional self-supervised learning such as DINOv2~\cite{oquab2023dinov2} and OpenCLIP~\cite{cherti2023reproducible} in image classification and semantic segmentation. These studies have shown that it is possible to build high performance recognition models while training only synthetic data and keeping data collection costs low by learning visual representations.
Inspired by these studies, we utilize data generated by generative models for contrastive learning.

\section{Problem Setting and Preliminary}
Inspired by the effectiveness of synthetic data as training data in 2D, we propose leveraging synthetic data as a solution to the scarcity of 3D data by using it to augment datasets. Specifically, we propose the method TeGA, that utilizes generative text-to-3D models to create datasets for language-image-3D pretraining.
Therefore, in this section, 
we introduce the proposed TeGA as a synthetic dataset expansion method, capable of learning rich geometric representations to enhance zero-shot recognition model generalization. 
Finally, we outline the formulation of a language-image-3D learning method, aiming to learn feature spaces for embedding alignment across three different modalities.

\noindent{\textbf{Problem Setting.}}
Zero-shot tasks generally demand large datasets to generalize to unseen classes~\cite{radford2021learning}, yet, in zero-shot 3D classification, the cost of collecting and annotating 3D data is a significant bottleneck and the performance of the models remains low~\cite{liu2024openshape, zhou2023uni3d, gao2024sculpting}. 
Recently, synthetic dataset expansion has achieved notable performance improvements in several fields~\cite{isola2022when, sariyildiz2023fake}.
Moreover, recent approaches in zero-shot 3D classification have demonstrated that pretraining on language-image-3D data enables knowledge distillation from models pretrained on other modalities, thereby improving performance under limited data conditions~\cite{qi2023contrast, hegde2023clip, huang2023clip2point, xue2024ulip, liu2024openshape}.
We consider these findings critical for improving the accuracy of zero-shot 3D classification. Hence, we present a multi-modal synthetic dataset expansion method called TeGA that generates synthetic language-image-3D data from text prompts using text-to-3D models. 

To frame the zero-shot 3D classification, we provide an overview of the dataset setup. We begin by training a model $f(\theta)$ to handle multiple types of visual inputs from a source dataset $D_{o} = \{(x_i^I, x_i^T, x_i^P)\}_{i=1}^{n_o}$, which consists of image ($x_{i}^I$), text ($x_{i}^T$), and 3D point cloud ($x_{i}^P$) modalities. The goal of this task is to classify $N$ semantic classes in a target dataset $D_{t} = \{(x_j^I, x_j^T, x_j^P)\}_{j=1}^{n_t}$.
Here, $n_o$ and $n_t$ indicate the number of samples, emphasizing the model's ability to generalize to previously unknown classes.

The core of dataset expansion lies in extending the source dataset with a synthetic dataset, where $D_{s} = \{(x_k^I, x_k^T, x_k^P)\}_{k=1}^{n_s}$, to enhance the model $f(\theta)$'s ability to generalize on unseen category shapes. The key to the task is to design $D_{s}$ to strengthen the adaptability of $f(\theta)$ to new concepts. The construction of this extended dataset, $D_{o} \cup D_{s}$, enriches the model's ability to recognize unseen category shapes.

\noindent{\textbf{Text-to-3D Models as Dataset Generator.}}
Cognitive science suggests that people use past experience to recognize unfamiliar objects~\cite{bar2004visual, kiani2007object}. For example, children use memories of other toys to imagine new ways to play with a new toy. 
This is a fundamental insight for the use of generative models as training data. The use of synthetic data generated from generative models for zero-shot recognition is very similar to the process by which humans recognize new concepts of objects from previous knowledge.
Recent advancements in text-to-3D models enable the generation of realistic 3D synthetic data that is both rich and semantically meaningful~\cite{nichol2022point, poole2022dreamfusion, lin2023magic3d, chen2023fantasia3d}. Unlike traditional generative models such as GANs~\cite{goodfellow2020generative} and VAEs~\cite{kingma2013auto}, text-to-3D models accept direct text prompts, allowing for user-specified, scenario-driven 3D shape generation. Therefore, this paper aims to improve the generalization ability of the model $f(\theta)$ by constructing a synthetic dataset through a text-to-3D model for dataset expansion.


Inspired by the recent success of synthetic data training using generative models based on diffusion models, we adopt a diffusion-based generative model as our text-to-3D model. Generally, text-to-3D models based on diffusion models can be simply represented as follows:

\begin{equation}
\label{eq:gtheta}
P = G_{\theta} \bigl(T, \omega \bigr).
\end{equation}

where $G_{\theta} \bigl( \cdot \bigr)$ represents the generator that inputs text $T$ and outputs a point cloud $T$, and $\omega$ represents the guidance scale that is used to adjust how strongly the input text is reflected in the point cloud generation process.
In this paper, we utilize a generative text-to-3D model $G_{\theta}(\cdot)$ as generating a point cloud $x_i^P$ from a text $x_i^T$.

\begin{figure}[tb]
    \centering
    \includegraphics[width=\linewidth]{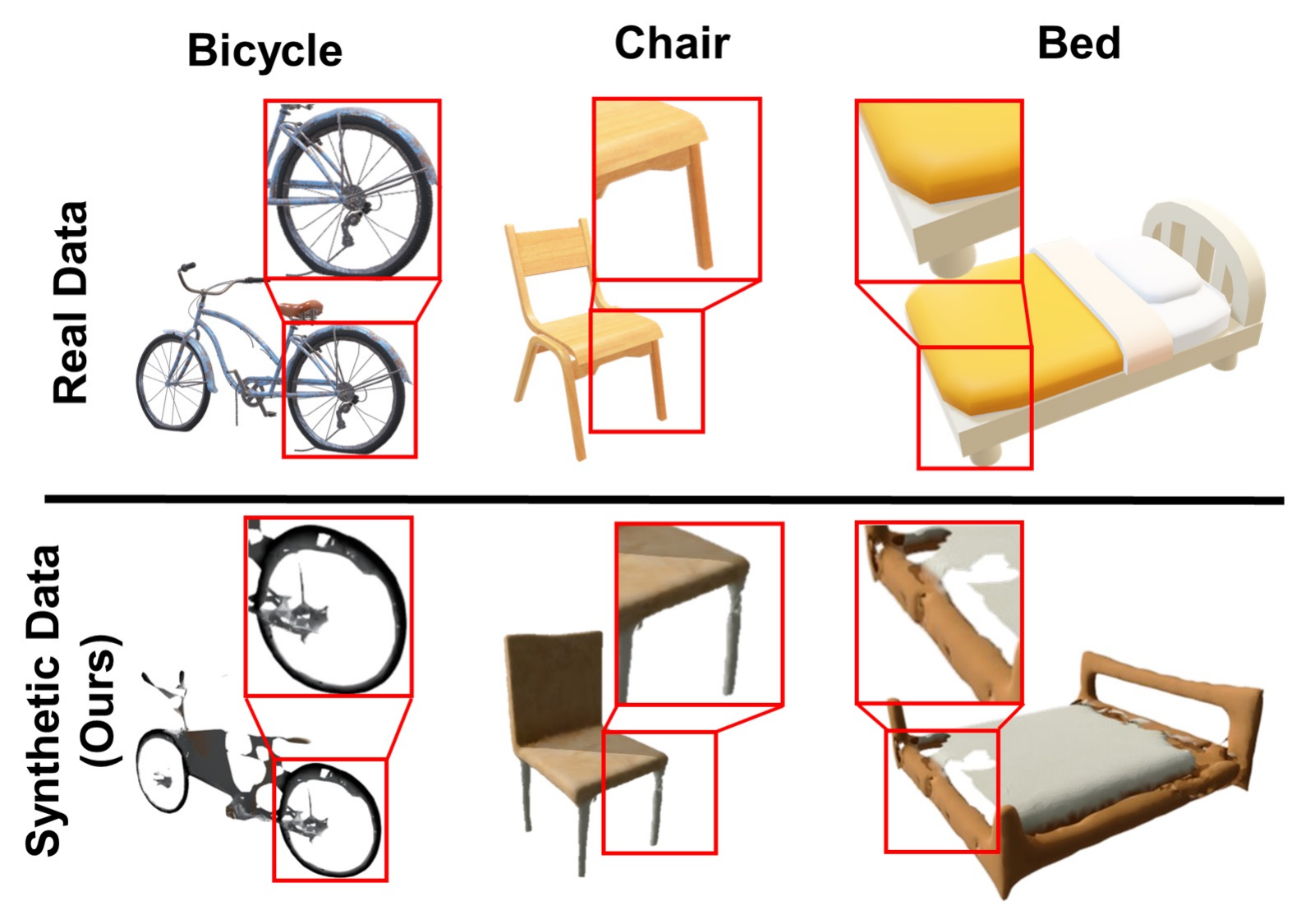}
    \caption{
    \kt{
    A visualization of synthetic 3D data generated from real 3D data and Point-E. Synthetic 3D data shows that it is more difficult to generate detailed geometrical detail compared to real data.
    }}
    \label{fig:enter-label_fig2}
    \vspace{-10pt}
\end{figure}

\noindent{\textbf{Language-Image-3D Contrastive Learning.}}
The goal of language-image-3D contrastive learning is to align the 3D embeddings with the rich, pre-trained feature spaces of images and text to facilitate the classification of 3D data including unseen classes. This facilitation enables language-image-3D contrastive learning to mitigate the impact of the 3D data scarcity problem. In fact, models leveraging language-image-3D contrastive learning outperform models trained with only text and 3D data in zero-shot 3D classification.
Given the typically limited 3D datasets, a practical approach is to leverage CLIP's knowledge as a shared embedding space. In this setup, we freeze CLIP's image and text encoders and align the 3D point cloud encoder to this shared space using contrastive learning. For a triplet of image, text, and point cloud $(x_i^I, x_i^T, x_i^P)$, the contrastive objective maximizes similarity within the shared embedding space as follows:
\begin{align}
L_{\text{All}} = -\frac{1}{2N} \sum_{i=1}^N \sum_{(A, B) \in \mathcal{S}} &\left( \log \frac{\exp(h_i^A \cdot h_i^B / \tau)}{\sum_j \exp(h_i^A \cdot h_j^B / \tau)} \right. \nonumber \\
&\left. + \log \frac{\exp(h_i^B \cdot h_i^A / \tau)}{\sum_j \exp(h_i^B \cdot h_j^A / \tau)} \right)
\end{align}
where $\mathcal{S}=\{(I,T), (P,I), (P,T)\}$ represents the pairs across modalities; in other words, $(I,T)$ denotes the image-text pair, $(P,I)$ the point cloud-image pair, and $(P,T)$ the point cloud-text pair. In addition, the normalized features are defined as $h_i^A = g_A(f_A(x_i^A)) / \| g_A(f_A(x_i^A)) \|$, where $f_A$ and $g_A$ denote the encoder and learnable projection head for each modality $A$. Similarly, $h_j^B$ is defined for modality $B$, with $(A, B) \in \mathcal{S}$ representing pairs across image, text, and point cloud modalities. 
The temperature parameter $\tau$ is a learnable parameter. It controls the strength of the penalty for samples with high similarity. Through this contrastive objective, we enable 3D shapes to align within CLIP's embedding space, facilitating a coherent language-image-3D representation.

\begin{figure}[tb]
    \centering
    \includegraphics[width=\linewidth]{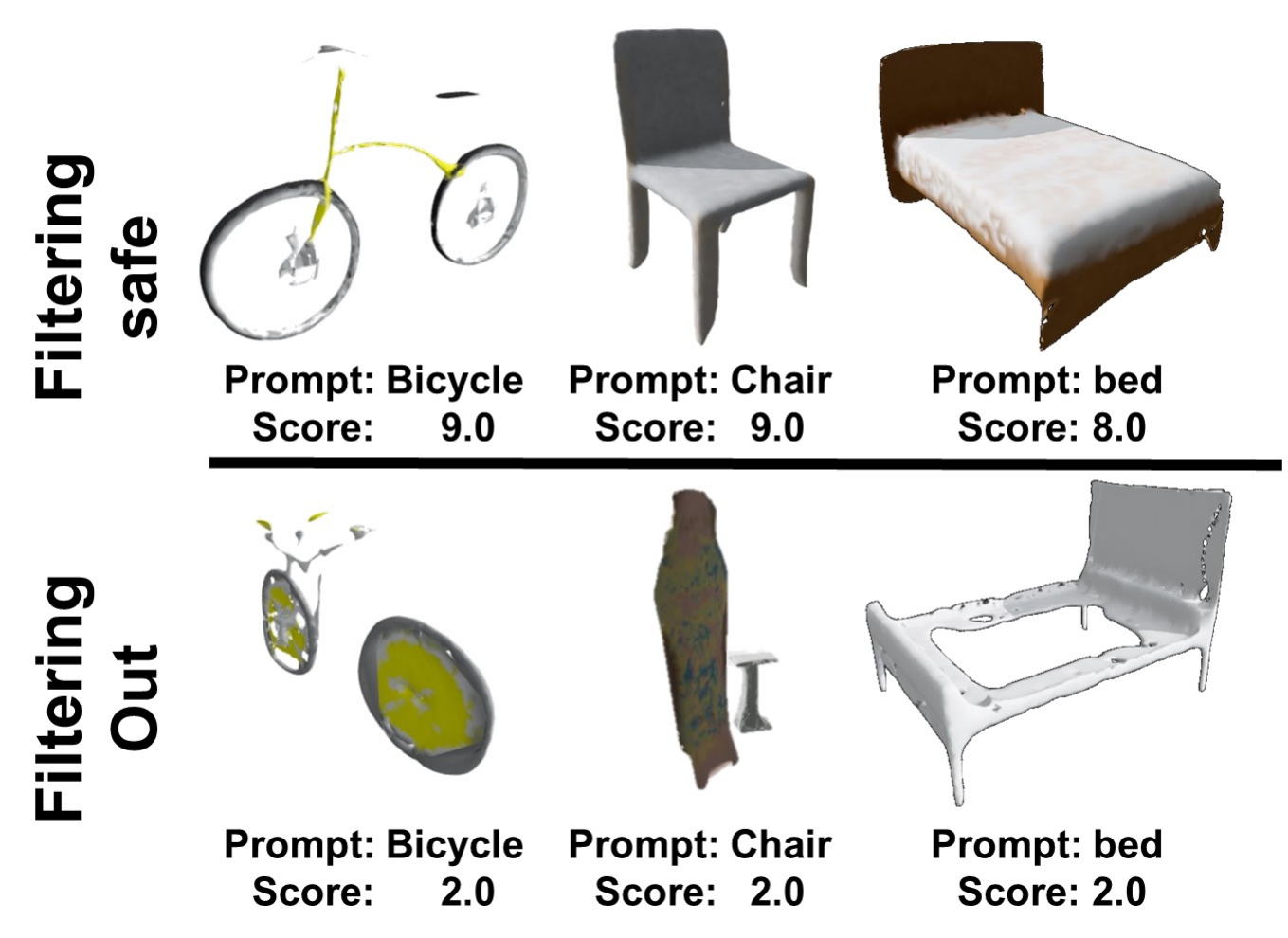}
    \caption{
    \kt{\
    A visualization of consistency filtering. The upper shows samples which passed filter; the lower shows samples which filtered out. 
    Our filtering can detect error cases while generation process.
    }}
    \label{fig:filtering_result}
    \vspace{-10pt}
\end{figure}
\begin{figure*}[t]
    \centering
    \includegraphics[width=\linewidth]{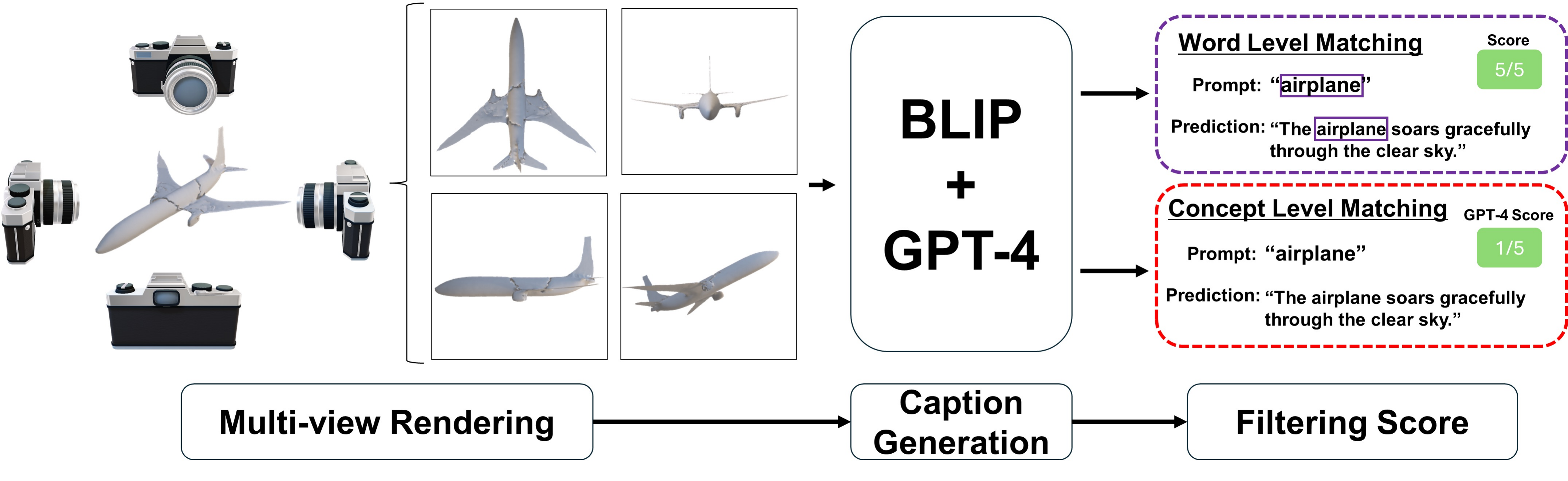}
    \caption{The overview of consistency filtering process.
    The purpose of this process is to remove misaligned data that may introduce model collapse during training. Specifically, rendered multi-view images are input into BLIP to generate captions. Then, the captions are summarized to one caption by GPT-4. Finally, the quality of the generated data is evaluated by comparing the text used for generation with the generated captions through two matching methods: word-level matching and concept-level matching.
    }
    \label{fig:enter-label_fig4}
    \vspace{-10pt}
\end{figure*}

\section{Proposed Method: TeGA}
In this section, we introduce TeGA, expanding language-image-3D datasets with high-quality synthetic 3D data generated from text-to-3D models.

\noindent{\textbf{Overview of TeGA.}}
To tackle the problem of 3D data scarcity in zero-shot 3D classification, the proposed method automatically generates a synthetic dataset composing language, image, and 3D modalities using a text-to-3D model.
In detail, to compose an expansion dataset, the proposed method combines the desired text, the point cloud generated by the text-to-3D model, and the images rendered from the generated point cloud.
The proposed method can generate large amounts of data without requiring human data collection or annotation and expand real datasets.
However, the generation process may err in some cases due to failure to generate or render, meaning that the alignment between each modality may not be accurate. This error may cause the model to collapse during training. 
To address this problem, we introduce a consistency filtering strategy of TeGA and the quality of the expanded dataset is improved by filtering low-quality data.
 
\noindent{\textbf{Synthetic Dataset Construction.}}
Our goal is to generate a synthetic dataset $D_{t} = \{(x_{i}^{I}, x0_{i}^{T \prime}, x_{i}^{P \prime})\}_{i=1}^{n_{o}} $ consisting of text $x_{i}^{T}$, images $x_{i}^{I \prime}$, and point clouds $x_{i}^{P \prime}$ using the text $x_{i}^{T}$ from the real dataset $D_{s} = \{(x_{i}^{I}, x_{i}^{T}, x_{i}^{P})\}_{i=1}^{n_{s}}$ that also comprises text $x_{i}^{T}$, images $x_{i}^{I}$, and point clouds $x_{i}^{P}$.
We use Point-E~\cite{nichol2022point} as the text-to-3D model of TeGA because Point-E is suitable for generating large amounts of data due to its ability to rapidly generate point clouds.
Also, we utilized only ShapeNet as the real dataset before applying TeGA instead of several datasets because our goal is not to outperform state-of-the-art methods but to serve as a training dataset expansion.
ShapeNet is a dataset containing 52,470 point cloud samples across 55 classes.


We first generate a point cloud by inputting the category names $x_{i}^{T}$ of the real dataset $D_s$ as text prompts into Point-E.
The generation process is represented as follows:
\begin{equation}
    x_{i}^{P \prime} = G_\theta \left( x_{i}^{T}, \omega \right)
\end{equation}
We have the point clouds and text, so we render images from the point clouds. However, without mesh information, the rendered images may appear to lack accurate shape details. To address this, we perform meshing using the Ball Pivoting Algorithm~\cite{bernardini1999ball}.
While various meshing techniques exist, Point-E's point clouds are not designed for meshing, and, in our experience, using state-of-the-art meshing methods often leads to the collapse of the rendered 3D data. Therefore, we adopted the older meshing technique, the Ball Pivoting Algorithm.
Then, images $x_{i}^{I \prime}$ are generated by rendering the 3D data from 20 viewpoints.
The viewpoints are determined by rotating it clockwise in 18-degree increments, with focal lengths dynamically determined by the Open3D library.
The processes of meshing and rendering are represented as follows:
\begin{equation}
    x_{i}^{I \prime}= \mathcal{R}(\mathcal{M}(x_{i}^{P \prime}))
\end{equation}
where $\mathcal{M}(\cdot)$ denotes meshing process and $\mathcal{R}(\cdot)$ denotes rendering process.
The sets of inputted text $x_{i}^{T}$, rendered images $x_{i}^{I \prime}$, and generated point clouds $x_{i}^{P \prime}$ that pass the consistency filtering are adopted as the synthetic dataset $D_t$.

Figure~\ref{fig:enter-label_fig2} shows examples of synthetic and real 3D data generated by Point-E. The real 3D data is a sample from Objaverse-LVIS. Figure~\ref{fig:enter-label_fig2} shows that the synthetic 3D data generated by Point-E outputs a 3D shape that matches the input text. On the other hand, there are cases where even the smallest details of the 3D shape are not generated accurately. For example, the synthetic 3D data generated by Point-E has difficulty in generating even the spokes of a bicycle.

\noindent\textbf{Consistency Filtering.}
Alignment across modalities is crucial for language-image-3D pretraining. 
In our generation process, alignment issues may occur at two stages: generating point clouds from text using Point-E and rendering images from point clouds. Failures in point cloud generation often result from Point-E's misinterpretation of the text. Also, rendering failures occur when the point cloud is unsuitable for meshing, such as when parts of the point cloud are incomplete. 
Training a model with such misaligned data could lead to a breakdown of modality alignment within the model. To address this, we apply consistency filtering to each generated data sample.


If alignment is maintained between the input text $x_{i}^{T}$ and the final output image $x_{i}^{I \prime}$ in the data generation process, it can be inferred that the intermediate output, the point cloud $x_{i}^{P \prime}$, is also aligned. This ensures that the alignment of the entire generated dataset is verified. Therefore, we evaluate the alignment between the text and images. Based on existing evaluation methods for 3D data~\cite{he2023t}, our filtering compares the integrated text generated from captions of multi-view images with the input text using GPT-4 and algorithm-based approaches. The process is shown in Figure~\ref{fig:enter-label_fig4}.

For each synthetic data, we first select two images showing the front and back of the 3D data among the generated images during the dataset creation process. 
Next, we input the two images into BLIP~\cite{li2023blip} to generate each caption.
To simplify, the generated captions for all viewpoints are combined into one unified text  $y_i$ using GPT-4.
The combined caption is then compared with the text used for generation and evaluates whether the synthetic data is aligned or not. The evaluation is conducted using two metrics: a text-based consistency metric and a semantic alignment metric. 
For the text-based consistency metric, we then compute a match score between this shared caption and the original text query. 
Specifically, if the generated caption $y_i$ contains the text $x_{i}^{T}$, a score of 5 is assigned; if it does not, a score of 1 is assigned.
\begin{equation}
\label{eq:text-based-consistency}
\text{s}_{\text{text}}(y_i, x_{i}^{T})
=
\begin{cases}
5, & \text{if } x_{i}^{T} \subseteq y_i, \\
1, & \text{otherwise}.
\end{cases}
\end{equation}
For the semantic alignment metric, inspired by the alignment evaluation in T3Bench~\cite{he2023t}, we use GPT-4 to evaluate the semantic alignment of the generated caption relative to the original prompt. 
Specifically, the common captions and text prompts generated from the two images are used to generate a five-point semantic-based similarity score using GPT-4. 
\begin{equation}
\label{eq:semantic-alignment}
\text{s}_{\text{sem}}\bigl(y_i, x_{i}^{T}\bigr)
=
\mathrm{GPT4}\bigl(y_i, x_{i}^{T}\bigr)
\;\in\;\{1, 2, 3, 4, 5\}.
\end{equation}
For more information about the GPT-4 prompts, see Appendix~\ref{app:details_tega}.

The two calculated scores $\text{s}_{\text{text}}, \text{s}_{\text{sem}}$ are summed to obtain the final score $s$. 
\begin{equation}
\label{eq:semantic-alignment}
\text{s}\bigl(y_i, x_{i}^{T}\bigr)
=
\text{s}_{\text{text}}\bigl(y_i, x_{i}^{T}\bigr) + \text{s}_{\text{sem}}\bigl(y_i, x_{i}^{T}\bigr)
\end{equation}
If this score $\text{s}\bigl(y_i, x_{i}^{T}\bigr)$ exceeds the threshold $\delta$, the data is considered aligned and included in the dataset; otherwise, it is discarded. 
We experimentally validated this threshold $\delta$ during data generation with ShapeNet and finally set it to 3.5 for use in the dataset. Filtering with this threshold in ShapeNet resulted in approximately half of the data being filtered out.
Figure~\ref{fig:filtering_result} shows samples of our consistency filtering.
The filtered data has well-formed meshes, making it visually easy to identify the objects. In contrast, the data that failed the filtering process often has collapsed meshes or represents unrecognizable objects.

\begin{table}[t]
  \centering
  \setlength{\tabcolsep}{7pt}
  \caption{Accuracy with and without filtering.}
  \vspace{-10pt}
  \resizebox{0.99\linewidth}{!}{
  \begin{tabular}{c|cccccccc}
    \toprule[0.8pt]
    \multirow{2}*{Filtering} & O-LVIS & S-Object & M40 \\
    & Top1 & Top1 & Top1 \\
    \midrule[0.5pt]
    w / o  filtering  & 11.1 & 45.4 & \textbf{73.4} \\
    w / filtering & \textbf{11.5} & \textbf{48.7} & 71.3 \\
    \bottomrule[0.8pt]
  \end{tabular}}
  \label{tab:noise_filtering}
  \vspace{-10pt}
\end{table}


\section{Experiments}
In this section, we evaluate the effectiveness of the proposed TeGA in zero-shot 3D classification. Section~\ref{sec:experimental_setup} first introduces our experimental setup, followed by Section~\ref{sec:exploratory_experiments}, which presents the results of the exploratory experiments. Additionally, Section~\ref{sec:comparison} discusses the main experimental results in comparison with our baseline and previous zero-shot 3D classification models. Finally, Section~\ref{sec:ablation_study} analyzes the key components of TeGA.

\subsection{Experimental Setup} 
\label{sec:experimental_setup} 
\noindent{\textbf{Training Dataset.}} 
Recent research has achieved high performance in zero-shot 3D classification using a 3D dataset that integrates four sources: ShapeNet~\cite{chang2015shapenet}, 3D-FUTURE~\cite{fu20213d}, ABO~\cite{collins2022abo}, and Objaverse~\cite{deitke2023objaverse}. 
However, our main goal is not to outperform state-of-the-art methods, but to test whether synthetic 3D data generated from text-to-3D models can serve as a training dataset expansion. For this reason, we use ShapeNet as a baseline in this paper.

\noindent{\textbf{Zero-Shot 3D Classification.}} 
Zero-shot 3D classification is the task of classifying objects of unseen categories that are not included in the training data. Following the experimental setup of MixCon3D, we use ModelNet40~\cite{wu20153d}, ScanObjectNN~\cite{uy2019revisiting}, and Objaverse-LVIS~\cite{deitke2023objaverse} as evaluation 3D datasets. ModelNet40 is a CAD dataset containing 12,311 samples of 40 categories, including chairs and tables. ScanObjectNN is captured from real-world environments using RGBD sensors and contains 2,902 samples of 15 categories. Objaverse-LVIS is a relatively large dataset with a collection of 1,156 categories, comprising a total of 46,206 samples.

\noindent{\textbf{Implementation Details.}}
We used eight Nvidia H100 GPUs with a batch size of 1,024. Other settings followed the default configurations of MixCon3D. We also employed PointBERT as the point cloud encoder in MixCon3D. 
We trained the model for 200 epochs with the AdamW optimizer~\cite{loshchilov2017decoupled}, a warmup epoch of 10, and a cosine learning rate decay schedule~\cite{loshchilov2016sgdr}. 
The base learning rate was set to 1e-3, based on the linear learning rate scaling rule: $lr=base\_lr \times \text{batchsize}/256$. 
The Text Encoder and the Image Encoder are frozen using OpenCLIP ViT-bigG-14~\cite{cherti2023reproducible}, and, as in Liu et al.~\cite{liu2024openshape}, simple layers are added afterward to optimize the model.
For dataset generation, we used five Nvidia TITAN RTX GPUs. The generation parameters followed the default settings of Point-E: the output point cloud size was 4,096, the guidance scale $\omega$ was set to [3.0, 0.0], and 50 diffusion steps were performed to obtain the final output.

\begin{table}[t]
  \centering
  \setlength{\tabcolsep}{7pt}
  \caption{Accuracy when varying Guidance scale.}
  \vspace{-10pt}
  \resizebox{0.85\linewidth}{!}{
  \begin{tabular}{c|cccccccc}
    \toprule
    \multirow{2}*{Guidance scale} & O-LVIS & S-Object & M40 \\
    & Top1 & Top1 & Top1\\
    \midrule
    0.3 & 11.4 & 44.8 & 70.5 \\
    3.0 & \textbf{11.5} & 46.6 & \textbf{71.1} \\
    30 & 10.3 & \textbf{48.3} & 69.7 \\
    \bottomrule
  \end{tabular}}
  \label{tab:guidance_scale}
  \vspace{-10pt}
\end{table}

\subsection{Exploratory Experiments} 
\label{sec:exploratory_experiments}
\begin{table*}[t]
\scriptsize
  \setlength{\tabcolsep}{1.8pt}
  \centering
   \caption{
   Comparison with zero-shot 3D classification performance on three representative benchmark datasets. 
   Best scores at MixCon3D are shown in underlined bold.
     }
  \vspace{-10pt}
  \adjustbox{width=\textwidth}{
    \begin{tabular}{l|c|cccc|cccc|cccc}
    \toprule[0.8pt]
    \multirow{2}[1]{1.0cm}{Method} & \multirow{2}[1]{*}{\begin{tabular}[c]{@{}c}Training  data\end{tabular}}  & \multicolumn{4}{c|}{Objaverse-LVIS} & \multicolumn{4}{c|}{ScanObjectNN} & \multicolumn{4}{c}{ModelNet40} \\          
    &  & Top1  & Top1-C & Top3  & Top5  & Top1 & Top1-C & Top3  & Top5  & Top1 & Top1-C & Top3  & Top5 \\
    \midrule[0.5pt]
    MixCon3D & ShapeNet & 9.4 & 5.7 & 16.7 & 20.4 & 46.5 & 41.6 & 67.9 & 79.6 & 64.6 & 60.3 & 83.0 & 87.1 \\ 
    MixCon3D \textbf{(Ours)} & ShapeNet + TeGA & \textbf{12.4} & \textbf{10.8} & \textbf{22.0} & \textbf{27.4} & \textbf{51.1} & \textbf{49.5} & \textbf{69.7} & \textbf{80.2} & \textbf{73.3} & \textbf{68.7} & \textbf{88.8} & \textbf{93.6} \\
    \bottomrule[0.8pt]
    \end{tabular}}
  \label{tab:zero_shot_main_results}
  \vspace{-10pt}
\end{table*}

This section analyzes the effect of key parameters in TeGA. Specifically, we conducted two exploratory experiments: (i) consistency filtering and (ii) guidance scale on Point-E. For each exploratory experimental baseline, we evaluated the performance of MixCon3D with a combined dataset of 10,000 synthetic data samples and ShapeNet.

\noindent{\textbf{Consistency Filtering (see Table~\ref{tab:noise_filtering}).}} 
The consistency filtering of the proposed TeGA is an important process because it removes the unaligned data from the generated data that adversely affects training. To verify the effectiveness of TeGA, we conducted a comparative experiment under two scenarios: one with consistency filtering applied and one without it. For this experiment, we use two versions of our synthetic dataset: one filtered by TeGA and one unfiltered, and combine each with the ShapeNet dataset. Then, these combined datasets are used for training. 
In Table~\ref{tab:noise_filtering}, we compare the Top1 accuracy with and without consistency filtering on Objaverse-LVIS, ScanObjectNN, and ModelNet40. The results show that the concatenated dataset with consistency filtering outperforms the unfiltered one by 0.4\% points on Objaverse-LVIS and 3.3\% points on ScanObjectNN, while it decreases 2.1\% points on ModelNet40. These results indicate that the proposed consistency filtering of TeGA provides effective synthetic training for zero-shot 3D classification. 

\noindent{\textbf{Effect of Guidance Scale of Point-E (see Table~\ref{tab:guidance_scale}).}} 
The guidance scale of Point-E is a critical parameter that controls the fidelity of the 3D model to the input text prompt. With a high guidance scale, the 3D model adheres more closely to the text prompt, though this reduces the diversity of geometric shapes. 
Conversely, a low guidance scale results in a 3D model that is less aligned with the text prompt but exhibits greater geometric diversity. 
Referring to the fact that the default guidance scale in Point-E is set to 3.0, this experiment evaluates varying guidance scale in \{0.3, 3.0, 30\}.

In Table~\ref{tab:noise_filtering}, we compare the Top1 accuracy with each guidance scale on Objaverse-LVIS, ScanObjectNN, and ModelNet40. 
Experimental results show that the guidance scale of 3.0 gives the best performance. 
When the guidance scale is set to 0.3, the generated point cloud is expected to have minimal reliance on the input text, leading to weaker alignment between modalities. Conversely, at a guidance scale of 30, the point cloud strongly reflects the input text; however, since all samples are generated using the same text, this likely results in a loss of diversity. This trade-off between alignment and diversity suggests that the highest accuracy is achieved with the standard guidance scale of 3.0.

\subsection{Comparison with Conventional Methods} 
\label{sec:comparison} 
In Table~\ref{tab:zero_shot_main_results}, we compare the zero-shot 3D classification results of MixCon3D. One model is trained with both ShapeNet and the synthetic dataset generated by TeGA, and the other is solely trained with ShapeNet. We generate a synthetic dataset with the same number of samples as ShapeNet. 
As a result, our proposed method effectively doubles the amount of training data compared to the original ShapeNet.
Table~\ref{tab:zero_shot_main_results} shows that our approach outperforms the baseline MixCon3D, which was trained solely on the original ShapeNet, across all benchmark datasets and evaluation metrics. Specifically, our method demonstrates substantial improvements, achieving gains of 3.0\% on Objaverse-LVIS, 4.4\% on ScanObjectNN, and 8.8\% on ModelNet40 in zero-shot performance.
These results clearly demonstrate that TeGA is an effective method for dataset expansion in zero-shot 3D classification, even when using synthetic 3D data. Furthermore, the ability to scale the dataset while maintaining or improving model performance opens up new possibilities for addressing data scarcity in 3D vision tasks.

\begin{figure*}[t]
    \centering
    \includegraphics[width=0.85\linewidth]{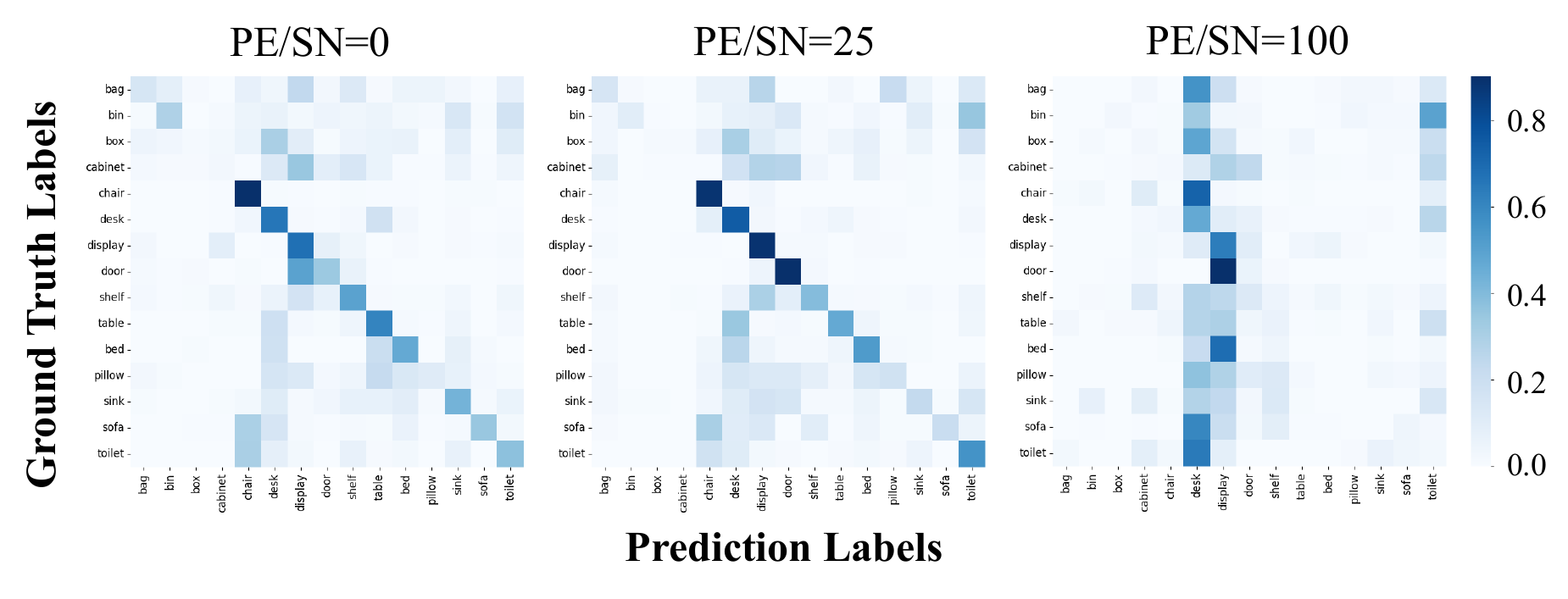}
    \vspace{-10pt}
    \caption{Confusion matrics when varying PE/SP. When  the proportion of real data decreases, the model's predictions increasingly skew towards desk and display, leading to an overall deterioration in prediction accuracy.}
    \label{fig:confusion_matrix}
    \vspace{-10pt}
\end{figure*}

\begin{table}[t]
  \centering
  \setlength{\tabcolsep}{7pt}
  \caption{Ablation studies of the balance of real and synthetic data. 
  We denote the replacement ratio of ShapeNet with synthetic data as Point-E (PE) / ShapeNet (SN)-\%.
  } 
  \vspace{-10pt}
  \resizebox{0.7\linewidth}{!}{
  \begin{tabular}{c|cccccccc}
    \toprule
    \multirow{2}*{PE/SN-\%} & O-LVIS & S-Object & M40 \\
    & Top1 & Top1 & Top1\\
    \midrule 
    0 & 9.01 & 50.3 & 63.7\\
    \midrule[0.5pt]
    25 & 9.47 & 47.4 & 67.7\\
    50 & 8.93 & 43.8 & 67.6 \\
    100 & 0.3 & 14.0 & 9.4 \\
    \bottomrule
  \end{tabular}}
  \label{tab:real_synthetic_ratio}
  \vspace{-10pt}
\end{table}

\subsection{Ablation Studies} 
\label{sec:ablation_study} 
This section analyzes the factors that contribute to performance improvements in dataset expansion using TeGA. Specifically, we investigate (i) the mixing ratio of synthetic 3D data and (ii) the scalability of synthetic 3D data.

\noindent{\textbf{Mixing ratio of synthetic 3D data (see Table~\ref{tab:real_synthetic_ratio} and Figure~\ref{fig:confusion_matrix}).}} In this experiment, we investigate whether synthetic data can serve as a substitute for real data. While keeping the total number of samples the same as the original ShapeNet, we replace a portion of the real data with synthetic data generated by TeGA.
Table~\ref{tab:real_synthetic_ratio} presents the Top1 performance in zero-shot 3D classification with varying proportions of synthetic 3D data. Surprisingly, it shows that the best performance is achieved when 25\% of the original ShapeNet is replaced with synthetic 3D data, with a performance improvement over using ShapeNet alone. However, as the proportion of synthetic 3D data increases beyond 25\%, performance gradually deteriorates; when only synthetic data is used, training fails and classification performance significantly drops. This decline is likely due to the learning strategy of MixCon3D, which extracts knowledge from CLIP models, making it difficult for the model to learn from the synthetic 3D data generated by Point-E due to domain gaps.

We also analyze why the model fails to learn when trained solely on synthetic data. In this experiment, we construct a mixture matrix at PE / SN = \{0, 25, 100\}, where real 3D data is progressively replaced by synthetic 3D data, to examine how the model misclassifies. As shown in Figure 4, for PE / SN = \{0, 25, 100\}, the categories predicted by the model exhibit similar trends. However, when PE/SN = 100, i.e., when the model is trained on synthetic 3D data, the predicted labels are biased towards desk or display.




\begin{table}[t]
  \centering
  \setlength{\tabcolsep}{7pt}
  \caption{Ablation studies of varying scale of synthetic data.}  
  \vspace{-10pt}
  \resizebox{0.7\linewidth}{!}{
  \begin{tabular}{c|cccccccc}
    \toprule
    \multirow{2}*{Scaling} & O-LVIS & S-Object & M40 \\
    & Top1 & Top1 & Top1\\
    \midrule
     \texttimes 0.1 & 11.1 & 48.2 & 71.0\\
     \texttimes 1 & 11.2 & 47.1 & 73.4 \\
     \texttimes 2 & 12.1 & 45.1 & 73.8 \\
    \bottomrule
  \end{tabular}
  }
  \label{tab:scaling_data}
  \vspace{-10pt}
\end{table}

\noindent{\textbf{Scalability of synthetic 3D data.}} 
In this experiment, we examine how scaling the amount of synthetic data added to ShapeNet influences the performance of MixCon3D. Specifically, we assess performance changes when Point-E generated data is scaled to 0.1x, 1x and 2x the original ShapeNet data. The scaling is based on ShapeNet's default dataset size of 53,470.
Table~\ref{tab:scaling_data} presents the Top1 performance with varying the amount of generated data. It shows that scaling plays a critical role in MixCon3D's training and that synthetic data effectively contributes to this scaling. According to these results, accuracy improves with the addition of scaled synthetic data for Objaverse-LVIS and ModelNet40. In contrast, the accuracy decreases with scaling for ScanObjectNN. This decline is likely due to ScanObjectNN's scores being more sensitive to noise.

\section{Conclusion}

To fix the problem of 3D data scarcity in zero-shot 3D classification, we propose TeGA (Text-guided Geometric Augmentation), a language-image-3D dataset expansion method with high-quality synthetic 3D data. TeGA plays a significant role in the expansion of existing datasets.
To further enhance the quality of 3D data, we apply consistency filtering to remove unaligned data that could negatively impact training. In practice, by combining the generated synthetic data with existing 3D datasets, we demonstrated improved performance in zero-shot 3D classification. 
Based on these results, we believe text-to-3D models have the potential to alleviate challenges associated with 3D data collection and, when leveraged for data expansion, can support the development of generalizable vision models for 3D vision.

\noindent{\textbf{Limitations.}} In this paper, we adopted Point-E as a text-to-3D model. However, Point-E is a generative model trained on data that is not open to the public. Since it is difficult to generate out-of-distribution data not included in the training dataset, our framework heavily relies on the generative model. Therefore, we plan to focus on improving the text-to-3D model in the future.
Additionally, when incorporating synthetic 3D data generated by Point-E into the training dataset, there is concern that societal biases or other unintended elements may be inadvertently introduced. It is recommended to take these risks into account and incorporate the findings from this study into the model improvement process. Furthermore, we look forward to the development of clean Text-to-3D models, like Stable Diffusion, which are built on publicly available datasets.

{
    \small
    \bibliographystyle{ieeenat_fullname}
    \bibliography{main}
}

\clearpage

\appendix
\setcounter{section}{0}
\setcounter{table}{0}
\setcounter{figure}{0}
\renewcommand{\thesection}{\Alph{section}}
\renewcommand{\thetable}{\Alph{table}}
\renewcommand{\thefigure}{\Alph{figure}}

\section{Details of TeGA}
\subsection{Details of the generated data by TeGA}
We provide samples of data generated by TeGA (\cref{fig:samples_passed} and \cref{fig:samples_filtered}).
The generated data which pass consistency filtering are shown in \cref{fig:samples_passed} and the generated data which are filtered out are shown in \cref{fig:samples_filtered}.
\cref{tab:class_ratios} compares some of the prompts used in the generation with the percentage that passed through the filter.

Classes that pass consistency filtering in large numbers tend to have clear class names. 
On the other hand, those that frequently fail consistency filtering often have ambiguous class names or simple shapes that make correspondence difficult to capture. 
For example, the 'birdhouse' has no fixed shape and could be a mesh or a wooden box. 
In such cases, it is difficult to generate and evaluate.
Conversely, the 'bicycle' has fixed the shape which is a vehicle with two wheels, a frame, handlebars for steering, and pedals for propulsion.
Some classes, despite having clear class names and not being simple in shape, are still prone to failing consistency filtering. 
\begin{table}[h!]
    \centering
    \adjustbox{width=\linewidth}{
    \begin{tabular}{@{}lcc@{}}
        \toprule
        \textbf{Class Name} & \textbf{Count (Filtered/Generated)} & \textbf{Percentage (\%)} \\ 
        \midrule
        airplane    & 3777/4045   & 93.38\%         \\ 
        bicycle     & 449/831     & 54.03\%         \\ 
        birdhouse   & 0/73        & 0.00\%          \\ 
        bookshelf   & 59/452      & 13.05\%         \\ 
        bottle      & 433/498     & 86.95\%         \\ 
        chair       & 6473/6778       & 95.05\%         \\ 
        clock       & 44/651      & 6.76\%         \\ 
        display     & 8/1093      & 7.32\%         \\ 
        lamp        & 1052/2318   & 45.38\%         \\ 
        telephone   & 42/1089     & 3.86\%         \\ 
        \bottomrule
    \end{tabular}}
    \caption{ShapeNet class name and the percentage that passed the filtering of the generation. A high percentage of the class names that are ambiguous are caught in the filtering process.}
    \label{tab:class_ratios}
\end{table}

\subsection{Details of the consistency filtering by TeGA}
\label{app:details_tega}
In \cref{tab:prompt_examples_success} and \cref{tab:prompt_examples_failed_total_ok}, \cref{tab:prompt_examples_failed_total_ng}, we show an example of a part of TeGA using GPT-4.
This evaluation is instructed by pre-defined prompt.
The prompt is based on T3Bench~\cite{he2023t}.
In T3Bench, the prompt for generation assumes a concrete long sentence so there needs only to perform sense-level comparisons between caption and prompt.
However, in our evaluation, since the prompt for generation is the class name of ShapeNet, which is a word, T3Bench always returns low score.
For these reasons, it is necessary to account for conceptual similarities and we redesigned the prompt for evaluation.

In the generated captions, key information is sometimes placed at the end of the sentence, and low scores are output by GPT-4 in such cases.
For example, in the case of \cref{tab:prompt_examples_failed_total_ok}, despite the presence of the same classmate word 'sofa', a low score is output. In such cases, performing word-level matching outputs a high score and is able to complement GPT-4's shortcomings.

\addtocounter{table}{3}

\newcommand{\cmark}{\checkmark} 
\newcommand{\xmark}{\textbf{\ding{55}}} 

\begin{table}[t]
    \centering
    \adjustbox{width=0.8\linewidth}{
    \begin{tabular}{@{}ccc|c@{}}
        \toprule
        \multirow{2}*{Text-PC} & \multirow{2}*{Image-PC} & \multirow{2}*{Text-3D} & ModelNet-40 \\ 
        \cmidrule(l){4-4}
        & & & Top1 \\ 
        \midrule
        \cmark & \xmark & \xmark & 56.3 \\
        \xmark & \cmark & \xmark & 35.9 \\
        \xmark & \xmark & \cmark & 0.4 \\
        \xmark & \cmark & \cmark & 60.3 \\
        \cmark & \cmark & \xmark & 42.2 \\
        \cmark & \cmark & \cmark & \textbf{73.3} \\
        \bottomrule
    \end{tabular}}
    \caption{Comparison of the final accuracy achieved when MixCon3D's contrastive learning is limited.  A check mark indicates an active contrastive learning setting, while a cross mark denotes an inactive setting. }
    \label{tab:multi-modal}
\end{table}

\addtocounter{table}{-4}
\addtocounter{figure}{2}
\begin{figure}[ht!]
    \centering
    \includegraphics[width=\linewidth]{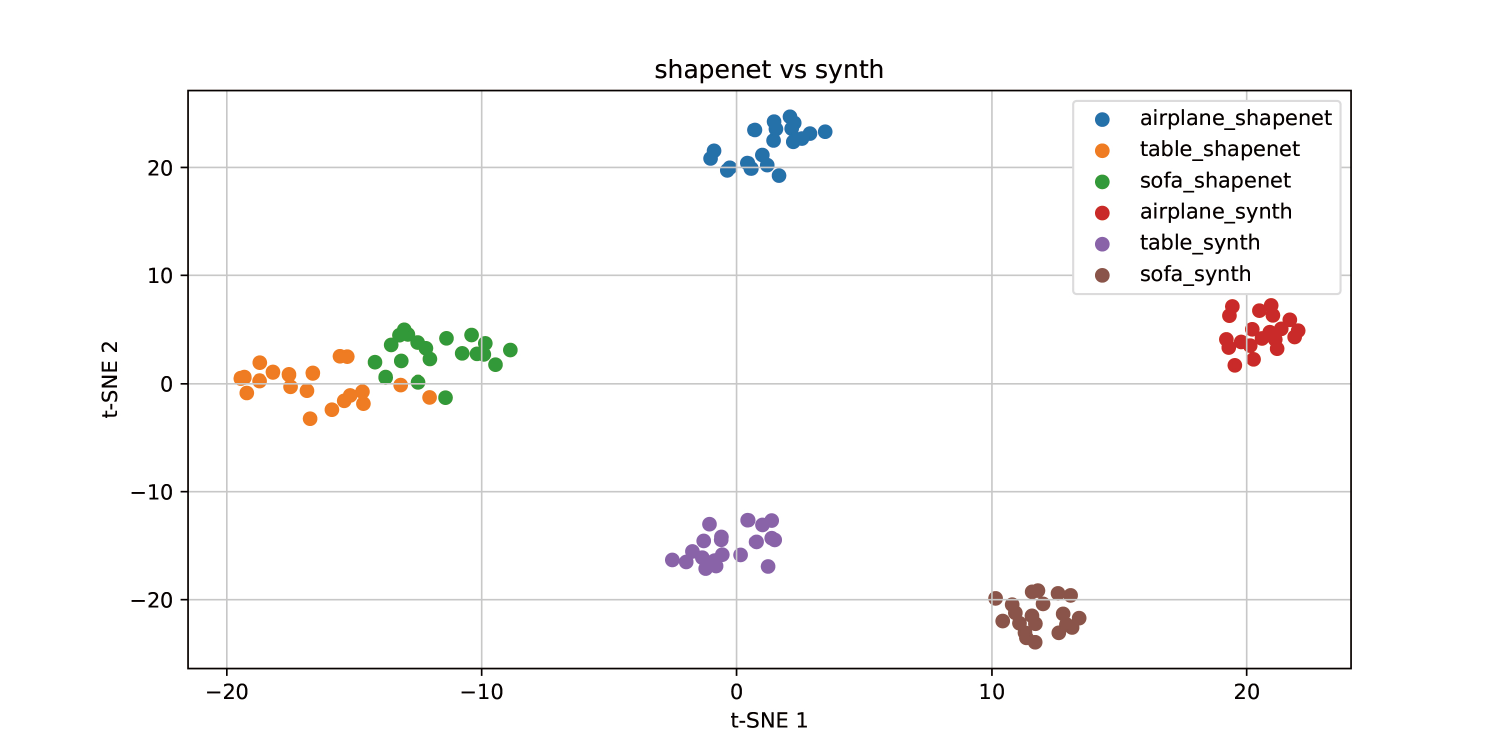}
    \caption{Visualization of features. Features with the same class name are clustered close to each other. However, real data and synthetic data are not clustered close to each other.}
    \label{fig:visualize_figure}
\end{figure}
\addtocounter{figure}{-3}

\section{Additional Experimental Results}
\noindent{\textbf{Multi-modal Pretraining.}}
We investigate which modality interactions contribute to training for language-image-3D tasks using synthetic data. 
Specifically, it compares the final accuracy achieved when MixCon3D's contrastive learning is applied to only a subset of the modalities. \cref{tab:multi-modal} presents the results. 
A check mark indicates an active contrastive learning setting, while a cross mark denotes an inactive setting. 
Note that the text-image contrastive learning is always active due to knowledge distillation from OpenCLIP~\cite{cherti2023reproducible}.

Surprisingly, high performance is achieved even without utilizing text-point cloud contrastive learning. Additionally, although training succeeds without text-3D contrastive learning, the accuracy is higher when it is included, indicating that text-3D contrastive learning definitely contributes to improving performance.

\begin{figure*}
    \centering
    \includegraphics[width=1.0\linewidth]{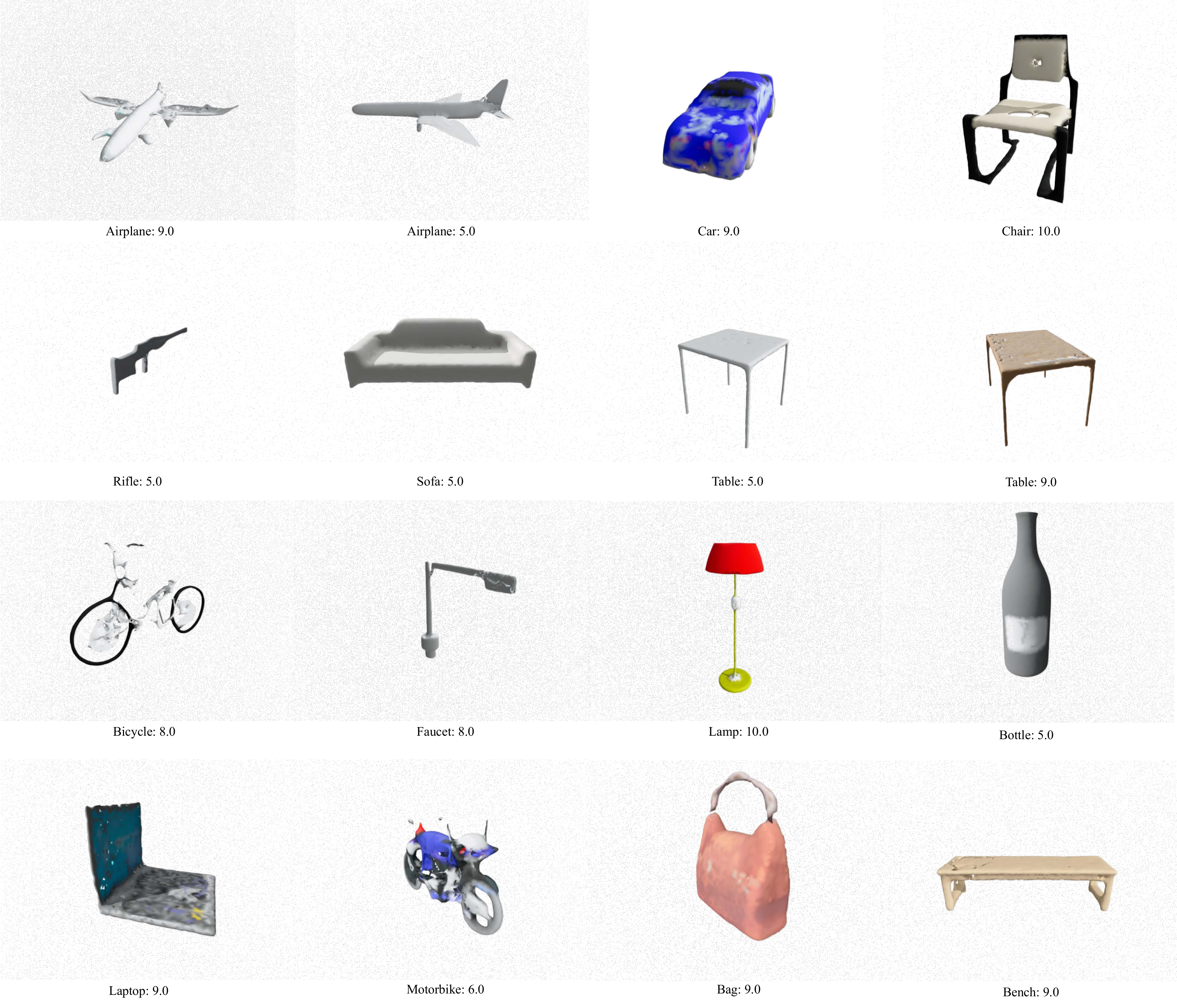}
    \caption{Samples of data generated by TeGA.}
    \label{fig:samples_passed}
\end{figure*}

\begin{figure*}
    \centering
    \includegraphics[width=1.0\linewidth]{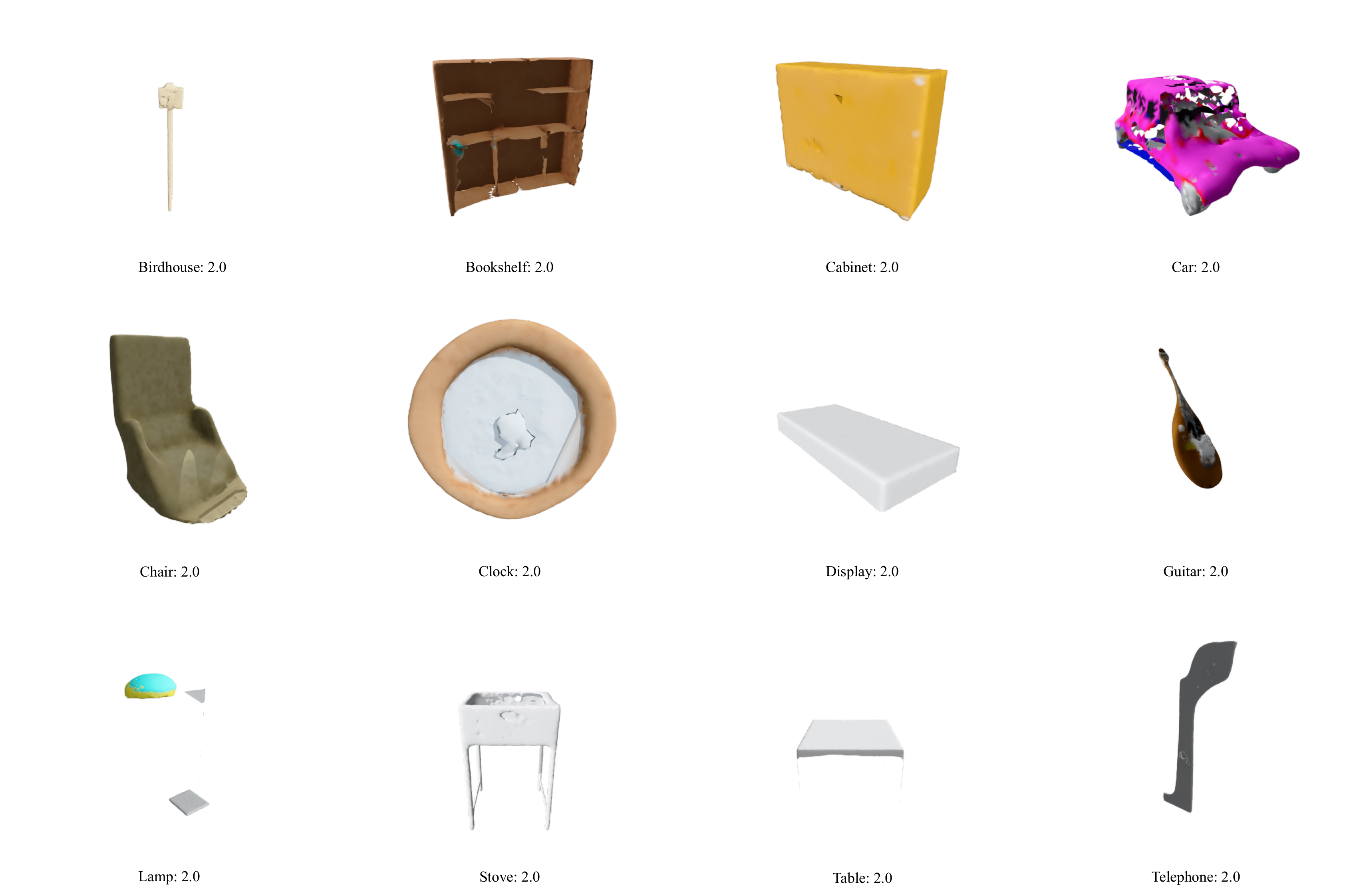}
    \caption{Samples of data generated but filtered out in TeGA.}
    \label{fig:samples_filtered}
\end{figure*}

\noindent{\textbf{Visualize Feature.}}
\cref{fig:visualize_figure} shows the visualization of the predicted feature by MixCon3D.
The features of the three classes—airplane, table, and sofa—are extracted from both ShapeNet data and synthetic data, and dimensionality reduction is performed using t-SNE.
Surprisingly, not only are different classes separated, but even the ShapeNet data and synthetic data of the same class are distinctly separated.

\begin{table*}[ht] 
    \centering
    \begin{tabular}{@{}p{2.5cm}p{12cm}@{}}
        \toprule
        \textbf{Prompt} & 
         You are an assessment expert responsible for prompt-prediction pairs. Your task is to score the prediction according to the following requirements:\newline
         1. Evaluate the recall, or how well the prediction covers the information in the prompt. If the prediction contains information that does not appear in the prompt, it should not be considered as bad.\newline
        2. Assign a score between 1 and 5, with 5 being the highest. Do not provide a complete answer; give the score in the format: 3\newline
        3. add points if the prediction and prompt are conceptually close (e.g. similar in appearance). (e.g., bike and bycicle and table and chair are close)\newline
        4. since the prompt is at the word level, it is inevitable that some detailed information is missing, so exclude it from the point deduction. \newline \newline
        prompt: car\newline
        prediction: A 3D rendering of a car with a pink and white exterior and a pink interior with red streaks\\
        \midrule
        \textbf{Answer:} &
        5 \\
        \midrule
        \textbf{Word matching} &
        5 \\
        \midrule
        \textbf{Total} &
        10 (safe) \\



        \bottomrule
    \end{tabular}
    \caption{The process of consistency filtering in TeGA. Captions and prompts are evaluated by GPT-4 and by word-matching. Both of evaluation are judged to be high.}
    \label{tab:prompt_examples_success}
\end{table*}
\begin{table*}[ht] 
    \centering
    \begin{tabular}{@{}p{2.5cm}p{12cm}@{}}
        \toprule
        \textbf{Prompt} & 
         You are an assessment expert responsible for prompt-prediction pairs. Your task is to score the prediction according to the following requirements:\newline
         1. Evaluate the recall, or how well the prediction covers the information in the prompt. If the prediction contains information that does not appear in the prompt, it should not be considered as bad.\newline
        2. Assign a score between 1 and 5, with 5 being the highest. Do not provide a complete answer; give the score in the format: 3\newline
        3. add points if the prediction and prompt are conceptually close (e.g. similar in appearance). (e.g., bike and bycicle and table and chair are close)\newline
        4. since the prompt is at the word level, it is inevitable that some detailed information is missing, so exclude it from the point deduction. \newline \newline
        prompt: sofa\newline
        prediction: A modern, cream-colored sofa\\
        \midrule
        \textbf{GPT-4 Answer} &
        1 \\
        \midrule
        \textbf{Word matching} &
        5 \\
        \midrule
        \textbf{Total} &
        6 (safe) \\



        \bottomrule
    \end{tabular}
    \caption{The process of consistency filtering in TeGA. Captions and prompts are evaluated by GPT-4 and by word-matching. The prompts appear in the caption, but are judged low by GPT-4, while word-matching is judged high. On the other hand, the word-matching score is high.}
    \label{tab:prompt_examples_failed_total_ok}
\end{table*}

\begin{table*}[ht] 
    \centering
    \begin{tabular}{@{}p{2.5cm}p{12cm}@{}}
        \toprule
        \textbf{Prompt} & 
         You are an assessment expert responsible for prompt-prediction pairs. Your task is to score the prediction according to the following requirements:\newline
         1. Evaluate the recall, or how well the prediction covers the information in the prompt. If the prediction contains information that does not appear in the prompt, it should not be considered as bad.\newline
        2. Assign a score between 1 and 5, with 5 being the highest. Do not provide a complete answer; give the score in the format: 3\newline
        3. add points if the prediction and prompt are conceptually close (e.g. similar in appearance). (e.g., bike and bycicle and table and chair are close)\newline
        4. since the prompt is at the word level, it is inevitable that some detailed information is missing, so exclude it from the point deduction. \newline \newline
        prompt: birdhouse\newline
        prediction: A black and white artistic object\\
        \midrule
        \textbf{GPT-4 Answer} &
        2 \\
        \midrule
        \textbf{Word matching} &
        1 \\
        \midrule
        \textbf{Total} &
        3 (ng) \\



        \bottomrule
    \end{tabular}
    \caption{The process of consistency filtering in TeGA. Captions and prompts are evaluated by GPT-4 and by word-matching. Both of evaluation are judged to be low.}
    \label{tab:prompt_examples_failed_total_ng}
\end{table*}


\end{document}